\documentclass[final,3p,times,twocolumn]{elsarticle}

\usepackage{graphics}
\usepackage{graphicx}
\usepackage{epsfig}

\usepackage{blindtext, graphicx, amsmath, algorithm, algpseudocode, pifont, algcompatible, comment, layout, amsthm, amssymb}
\usepackage{enumitem}   

\usepackage{float}

\usepackage[utf8]{inputenc}
\usepackage[english]{babel}

\makeatletter
\def\ps@pprintTitle{%
 \let\@oddhead\@empty
 \let\@evenhead\@empty
 \def\@oddfoot{}%
 \let\@evenfoot\@oddfoot}
\makeatother

\begin{document}

\begin{frontmatter}

\title{Offensive Language Detection: A Comparative Analysis}
\author{Vyshnav M T\corref{cor1}}
\author{Sachin Kumar S}
\author{Soman K P} 
\address{Center for Computational Engineering \& Networking (CEN)\\
Amrita School of Engineering, Coimbatore.\\
Amrita Vishwa Vidyapeetham, India.}
\cortext[cor1]{Corresponding author:}
\ead{vyshnav94.mec@gmail.com, kp\_soman@amrita.edu}

\begin{abstract}
Offensive behaviour has become pervasive in the Internet community. Individuals take the advantage of anonymity in the cyber world and indulge in offensive communications which they may not consider in the real life. Governments, online communities, companies etc are investing into prevention of offensive behaviour content in social media. One of the most effective solution for tacking this enigmatic problem is the use of computational techniques to identify offensive content and take action. The current work focuses on detecting offensive language in English tweets. The dataset used for the experiment is obtained from SemEval-2019 Task 6 on Identifying and Categorizing Offensive Language in Social Media (OffensEval). The dataset contains 14,460 annotated English tweets. The present paper provides a comparative analysis and Random kitchen sink (RKS) based approach for offensive language detection. We explore the effectiveness of Google sentence encoder, Fasttext, Dynamic mode decomposition (DMD) based features and Random kitchen sink (RKS) method for offensive language detection. From the experiments and evaluation we observed that RKS with fastetxt achieved competing results. The evaluation measures used are accuracy, precision, recall, f1-score.
\end{abstract}


\end{frontmatter}


\section{Introduction}
\label{sec1}
In this digital era, online discussions and interactions has become a vital part of daily life of which a huge part is covered by social media platforms like twitter, facebook, instagram etc. Similar to real life there exist anti-social elements in the cyberspace, who take advantage of the anonymous nature in cyber world and indulge in vulgar and offensive communications. This includes bullying, trolling, harassment \cite{1,2} and has become a growing concern for governments. Youth experiencing such victimization was recorded to have psychological symptoms of anxiety, depression, loneliness \cite{2}. Thus it is important to identify and remove such behaviours at the earliest. One solution to this is the automatic detection using machine learning algorithms.\\

Detecting offensive language from social media is a challenging research problem due to the different level of ambiguities present in the natural language and noisy nature of the social media language. Moreover, social media subscribers are from linguistically diverse and varying communities. Overseeing the complication of this problem, \cite{3} organized a task in SemEval2019, Task 6: Identifying and Categorizing Offensive Language in Social Media. The tweets were collected by the organizers using Twitter API and have annotated them in a hierarchical manner as offensive language present in the tweet, type of the offense and target of the offense. There were three sub-tasks according to the hierarchy of annotation: a) To detect if a post is offensive (OFF) or not (NOT), b) To Identify the type of offense in the post as targeted threat (TTH), targeted insult (TIN), untargeted (UNT), c) To identify if offense is targeted to organization or entity (ORG), group of people (GRP), individual (IND), or other (OTH). \\

The dataset had the following challenges:
\begin{itemize}
    \item Dataset was comparatively smaller.
    \item Dataset was biased/imbalanced \cite{4}.
\end{itemize}
 In this paper, we are proposing a comparative analysis for the sub-task A :Offensive language identification of the SemEval2019, Task 6. Sub-task A was the most popular sub-task among the three and had total 104 team participation. In Table \ref{tab:7}, the list of first 5 participants along with system and f1-score has been shown.\\
 
Offensive language detection is one of the challenging and interesting topic for research. Recent past had multiple shared tasks and research on this topic. One of the initial work on offensive language using supervised classification was done by Yin et al. \cite{9}. They have used Ngram, TFIDF and combination of TFIDF with Sentiment and Contextual as Features. Schmidt and Wiegand \cite{10} gave a survey on automatic hate speech detection using NLP. The authors surveyed on features like Simple Surface Features, Word Generalization, Linguistic Features, Knowledge-Based Features, Multimodal Information etc. In 2013, a study on detecting cyberbullying in YouTube comments was done by Dadvar et al. \cite{11}. They have used a combination of content-based, cyberbullying-specific and user-based features and showed that detection of cyberbullying can be improved by taking user context into account. Shared task GermEval 2018 organised by Wiegand et al.,\cite{12} was focused on offensive language detection on German tweets. It had a dataset of over 8,500 annotated tweets and was trained for binary classification of offensive and non-offensive tweets. They obtained an overall maco-F1 score of 76.77\%. Another shared task on Aggression Identification in Social Media was organised by Kumar et al., \cite{13}. The task provided a dataset with 15,000 annotated Facebook posts and comments in Hindi and English. They obtained a weighted F-score of 64\% for both English and Hindi. The rest of the paper is structured as follows. Section 2 explains about the methodology with formulation. Section 3 discusses on the Proposed approach. Section 4 talks on the Experiments and discussions performed. Finally, conclusion is given in Section 5.
\section{Methodology}
\label{sec2}
\subsection{Data Pre-processing}
Data pre-processing is a very crucial step which needs to be done before applying any machine learning tasks, because the real time data could be very noisy and unstructured.  For the two models used in this work, pre-processing of tweets is done separately:
\begin{itemize}
    \item \textbf{Pre-processing for Google model:} \\ It has become a common culture to use \#tags across social media. So we have replaced multiple \#tags with a single \#tag. Mostly @ symbol id used to mention person or entities in a tweet. So we replace multiple @symbols with a single @-mention. Some tweets may contain the link to a website or some other urls. So we replace all of these with a single keyword URLS.
   \item \textbf{Pre-processing for fasttext model:} \\  For applying fasttext model to get word vectors, we followed a different set of pre-processing steps. First, all the numbers, punctuation marks, urls (http:// or www.) and symbols (emoji, \#tags, \@-mention) were removed from the tweet as it do not contain information related to sentiment. After that, tokenization and lowercasing was applied to the tweets. Tokenization was done using tokenizer from NLTK package \cite{18}. Finally, the stop word are removed. The list is obtained from NLTK package. 
\end{itemize}

\subsection{Embeddings}
Word embeddings are ubiquitous for any NLP problem, as algorithms cannot process the plain text or strings in its raw form. Word emeddings are vectors that captures the semantic and contextual information of words. The word embedding used for this work are:
\begin{itemize}
    \item \textbf{FastText:} The fastText algorithm created by Facebook \cite{16} assumes every word to be n-grams of character. It helps to give the vector representations for out of vocabuary words. For the current work, fasttext based word embedding is used for generating token vectors of dimension 300 \cite{14}. Each vector corresponding to the tweet is generated by taking the average of token vectors.
    \item \textbf{Universal Sentence Encoder:} Developed by Google, Universal sentence encoder \cite{15,17} provides embeddings at sentence level. The dimension of the embedding vector is 512, irrespective of the number of tokens in the input tweet. These vectors can capture good semantic information from the sentences. For each tweet, this model generates a 512 length embedding vector and is used as features for further classification.
    \item \textbf{DMD and HODMD:} DMD is a method initially used in fluid dynamics which captures spatio-temporal features \cite{19}. It has been used in background-foreground separation \cite{20}, load forecasting \cite{21}, saliency detection \cite{22} etc. For natural language processing, DMD has been first applied for sentiment analysis \cite{23,24}. This motivated to explore DMD based feature for the present work.\\
    Dynamic mode decomposition (DMD) is a much more powerful concept and it assumes the evolution of the function over the rectangular field is effected by the mapping of a constant matrix $A$. $A$ captures the system’s inherent dynamics and the aim of the DMD is to understand using its dominant eignevalues and eigenvectors. Assumption is that this matrix $A$ is of low rank and hence the sequence of vectors $ \mathop {{x_1}}\limits_|^| ,\mathop {{x_2}}\limits_|^| ,\mathop {{x_3}}\limits_|^| ,...\mathop {,{x_k}}\limits_|^| ,...,\mathop {{x_{m + 1}}}\limits_|^| $ finally become a linearly dependent set. That is, vector $ \mathop {{x_{m + 1}}}\limits_|^|$  become linearly dependent on previous vectors.  The data matrix X in terms of eigen vectors associated with matrix $A$.
    \begin{equation}
       {A^k}{x_1} = \Phi {\Lambda ^k}{\Phi ^\dag }{x_1} = \Phi {\Lambda ^k}b
    \end{equation}
    \begin{equation}
       {A^k}{x_1} = \Phi {\Lambda ^k}{\Phi ^\dag }{x_1} = \Phi {\Lambda ^k}b = {x_{k + 1}}
    \end{equation}
    where, ${\Phi ^\dag }$ is pseudo inverse of ${\Phi }$. $A$ is of rank m and ${\Phi }$ have m columns. Hence, pseudo-inverse will do the job than inverse operation. The columns of ${\Phi }$ are called DMD modes and this forms the features.
    \begin{equation}
        \begin{array}{l}
    {x_{k + 1}} = \Phi \left( {\begin{array}{*{20}{c}}
    {\lambda _1^k{b_1}}\\
    {\lambda _2^k{b_2}}\\
     \vdots \\
    {\lambda _m^k{b_m}}
    \end{array}} \right)\\
    {\rm{      }} = \left( {\begin{array}{*{20}{c}}
    {\mathop {{\phi _1}}\limits_|^| }&{\mathop {{\phi _2}}\limits_|^| }& \ldots &{\mathop {{\phi _m}}\limits_|^| }
    \end{array}} \right)\left( {\begin{array}{*{20}{c}}
    {\lambda _1^k{b_1}}\\
    {\lambda _2^k{b_2}}\\
     \vdots \\
    {\lambda _m^k{b_m}}
    \end{array}} \right)
    \end{array}
    \end{equation}
    Time-lagged matrices are prepared as snapshot for this approach. In Eigensent \cite{25}, the authors proposed HODMD to find embedings for sentences. The authors suggested, sentences can be represented as a signal using word embeddings by taking the average of word vectors. This is intuitive because word embeddings almost obeys the laws of linear algebra, by capturing word analogies and relationships. Therefore, by considering every sentence as a multi-dimensional signal, we can capture the important transitional dynamics of sentences. Also, for the signal representation of sentences, each word vector will act as a single point in the signal. For the present work, to generate DMD and HODMD based features, Fastext based embedding is used.
\end{itemize}

\section{Proposed Approach}
\begin{figure}[H]
\centering
\includegraphics[height=6cm,width=3cm,clip]{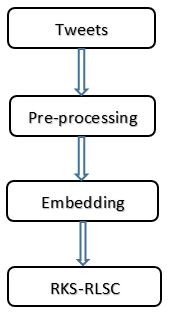}
\caption{Illustrates the block diagram for the proposed approach}
\label{fig:1}
\end{figure}

RKS approach proposed in \cite{26,27}, explicitly maps data vectors to a space where linear separation is possible. It has been explored for natural language processing tasks \cite{29,30}. The RKS method provides an approximate kernel function via explicit mapping.
\begin{equation}
\label{eq:10}
K({x_1},{x_2}) = \left\langle {\phi ({x_1}),\left. {\phi ({x_2})} \right\rangle } \right. \approx \left\langle {\left. {Z({x_1}),Z({x_2})} \right\rangle } \right.
\end{equation}
Here, $\phi(.)$ denotes the implicit mapping function (used to compute kernel matrix), $Z(.)$ denotes the explicit mapping function using RKS and ${\Omega _k}$ denotes random variable .
\begin{equation}
\label{eq:11}
Z(x) = \sqrt {1/k} \left[ {\begin{array}{*{20}{c}}
{Cos({x^T}{\Omega _1})}\\
 \vdots \\
{Cos({x^T}{\Omega _k})}\\
{Sin({x^T}{\Omega _1})}\\
 \vdots \\
{Sin({x^T}{\Omega _k})}
\end{array}} \right]\
\end{equation}
 Figure \ref{fig:1} show the block diagram of the proposed approach. 

\section{Experiments and Discussions} 
\label{sec3}
\subsection{Data Description}
OLID (Offensive Language Identification Dataset) is a collection of English tweets which are annotated using a three-level hierarchical annotation model. It was collected using Twitter API and contains 14,460 annotated tweets. The task divided the data as train, trial and test, of which train and trial was initially released as starting kit, finally test was released as Test A release. All three of these partitions were highly biased, thus making the task more challenging and real time. The train set had 13,240 tweets, out of which 8840 tweets were not offensive (NOT) and 4400 tweets were offensive (OFF). Similarly, test set had 860 tweets, which had 620 not offensive and 280 offensive tweets. Table\ref{tab:1} show the data distribution of the entire dataset. For the current work, train and test data are taken which is 14,100 tweets in number.
\begin{table}[H]
\centering
\begin{tabular}{|c|c|c|c|}
\hline
\begin{tabular}[c]{@{}c@{}}Target \\ Class\end{tabular} & Train data & \begin{tabular}[c]{@{}c@{}}Validation \\ data\end{tabular} & Test data \\ \hline
NOT & 8840 & 243 & 620 \\ \hline
OFF & 4400 & 77 & 280 \\ \hline
\end{tabular}
\caption{Illustrates the Data distribution}
\label{tab:1}
\end{table}
In Sub-task A: Offensive language identification, the goal was to discriminate offensive and not-offensive twitter posts. The target classes for each instance were a) Offensive (OFF): posts that contain any form of profanity or targeted offence. This includes threats, insults, and any form of untargeted profanity. b) Not Offensive (NOT): posts that doesn't have any profanity or offense in it. The result and discussion of the top 10 teams for the sub-task A is in section for Introduction. In that, team \cite{4} obtained highest f1-score of 82.9\%

\subsection{Results and Comparisons}
This section describes the result as three different cases. Case 1 \& 2 provides baseline approach to compare with the proposed RKS approcah described in case 3. Table \ref{tab:7} gives the results of the top 5 teams of sub-task A. Team with rank 1 achieved a maximum f1-score of 82.9\%.

\begin{table}[H]
\fontsize{9pt}{9pt}
\selectfont
\centering
\begin{tabular}{|c|c|c|c|}
\hline
\textbf{Rank} & \textbf{Team Name} & \textbf{System} & \textbf{F1} \\ \hline
1. & NULI \cite{4} & Bert & 0.829 \\ \hline
2. & \begin{tabular}[c]{@{}c@{}}Nikolov-\\ Radivchev \cite{5}\end{tabular} & Bert-Large & 0.815 \\ \hline
3. & UM-IU@LING \cite{6} & Bert-base-uncased & 0.814 \\ \hline
4. & Embeddia \cite{7} & Bert & 0.808 \\ \hline
5. & MIDAS \cite{8} & \begin{tabular}[c]{@{}c@{}}Bi -LSTM \\ with attention, \\ and\\  Bi - LSTM \\ + Bi - GRU\end{tabular} & 0.807 \\ \hline
\end{tabular}
\caption{Illustrates results of top 5 teams in semeval 2019: Task 6  sub-task A}
\label{tab:7}
\end{table}

\subsubsection{Case 1: Embeddings approach}
In this work, we have selected word vectors generated by Google universal encoder model, Fasttext, and DMD based features. The classification using the selected features are performed using machine learning algorithms such as Random Forest (RF), Decision Tree (DT), Naive Bayes (NB), Support vector machine (SVM) linear and RBF kernels,  Logistic Regression, and Random kitchen sinks. The evaluation measures used are accuracy (Acc.), precision (Prec), recall, f1-score (F1). Table \ref{tab:2} shows the classification result obtained for classical machine learning algorithms using the Google universal sentence encoder model features. It can be observed that svm linear classifier and Logistic regression has given maximum accuracy of 82.44\% and 82.56\%. \\
\begin{table}[H]
\fontsize{10pt}{10pt}
\selectfont
\centering
\begin{tabular}{|c|c|c|c|c|}
\hline
\multicolumn{5}{|c|}{\textbf{Google Embedding Vectors}} \\ \hline
\textbf{Algorithm} & \textbf{\begin{tabular}[c]{@{}c@{}}Acc.\\ (\%)\end{tabular}} & \textbf{\begin{tabular}[c]{@{}c@{}}Prec\\ (\%)\end{tabular}} & \textbf{\begin{tabular}[c]{@{}c@{}}Recall\\ (\%)\end{tabular}} & \textbf{\begin{tabular}[c]{@{}c@{}}F1\\ (\%)\end{tabular}} \\ \hline
\begin{tabular}[c]{@{}c@{}}RF\\ (n\_estm=30)\end{tabular} & 77.56 & 78.89 & 61.71 & 62.68 \\ \hline
NB & 74.19 & 68.44 & 69.58 & 68.92 \\ \hline
SVM Linear & 82.44 & 81.13 & 72.63 & 75.10 \\ \hline
SVM RBF & 72.09 & 36.05 & 50.00 & 41.89 \\ \hline
LR & 82.56 & 81.71 & 72.45 & 75.04 \\ \hline
\end{tabular}
\caption{Performance evaluation of Universal encoder model features using classical machine learning algorithms}
\label{tab:2}
\end{table}

\begin{table}[H]
\fontsize{10pt}{10pt}
\selectfont
\centering
\begin{tabular}{|c|c|c|c|c|}
\hline
\multicolumn{5}{|c|}{\textbf{Fasttext Embeddings}} \\ \hline
\textbf{Algorithm} & \textbf{\begin{tabular}[c]{@{}c@{}}Acc.\\ (\%)\end{tabular}} & \textbf{\begin{tabular}[c]{@{}c@{}}Prec.\\ (\%)\end{tabular}} & \textbf{\begin{tabular}[c]{@{}c@{}}Recall\\ (\%)\end{tabular}} & \textbf{\begin{tabular}[c]{@{}c@{}}F1\\ (\%)\end{tabular}} \\ \hline
\begin{tabular}[c]{@{}c@{}}RF\\ (n\_estm=30)\end{tabular} & 76.98 & 75.84 & 61.56 & 62.53 \\ \hline
NB & 52.33 & 58.40 & 59.78 & 51.89 \\ \hline
SVM Linear & 81.16 & 82.99 & 68.29 & 70.94 \\ \hline
SVM RBF & 74.42 & 79.88 & 54.68 & 51.38 \\ \hline
LR & 81.16 & 82.99 & 68.29 & 70.94 \\ \hline
\end{tabular}
\caption{Performance evaluation of Fasttext model features using classical machine learning algorithms}
\label{tab:3}
\end{table}
Table \ref{tab:3} shows the classification results obtained using the features generated by fasttext model for classical machine learning algorithms. For the fasttext model also, svm linear and logistic regression model have given maximum accuracies of 81.16\% respectively. \\

\subsubsection{Case 2: DMD approach}
\begin{figure}[H]
\centering
\includegraphics[height=3.5cm,width=7.8cm,]{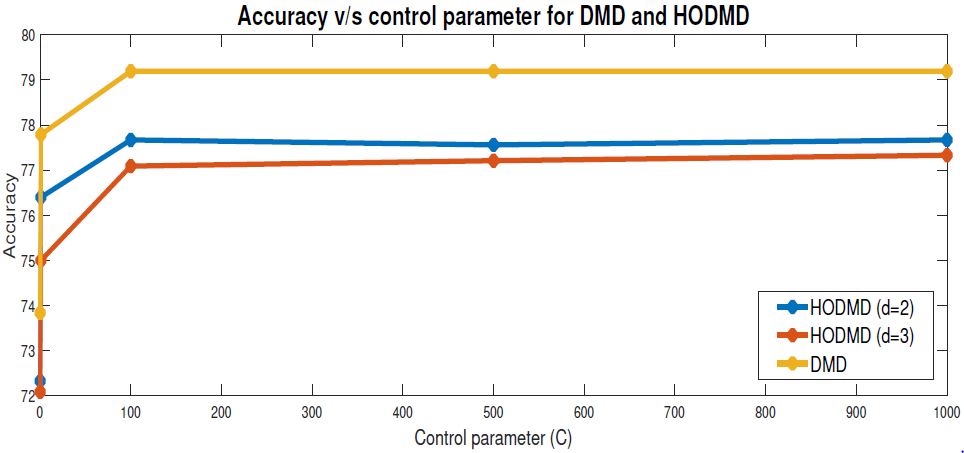}
\caption{Illustrates the Accuracy v/s control parameter for DMD and HODMD}
\label{fig:2}
\end{figure}

In order to provide a comparison, we explore DMD based features. The Table \ref{tab:4} shows the result obtained for normal DMD and HODMD based feature. The order for HODMD for the present work is 2 \& 3. The classification is performed using SVM-linear kernel with control parameter value chosen as 1000 as the suitable one. We tried for other values such as 0.1, 1, 100, 500, and 1000. Figure \ref{fig:2} shows the control parameter versus accuracy plot which helped to fix the parameter value.

\begin{table}[H]
\fontsize{10pt}{10pt}
\selectfont
\centering
\begin{tabular}{|c|l|l|l|}
\hline
\multicolumn{1}{|l|}{\textbf{}} & \multicolumn{2}{c|}{\textbf{HODMD}} & \multicolumn{1}{c|}{\textbf{DMD}} \\ \hline
\textbf{\begin{tabular}[c]{@{}c@{}}order $\rightarrow$\\  Score $\downarrow$\end{tabular}} & \multicolumn{1}{c|}{\textbf{d=2}} & \multicolumn{1}{c|}{\textbf{d=3}} & \textbf{} \\ \hline
\textbf{Acc. (\%)} & 77.67 & 77.33 & 79.19 \\ \hline
\textbf{Prec. (\%)} & 74.59 & 74.18 & 76.63 \\ \hline
\textbf{Recall (\%)} & 64.47 & 63.72 & 67.31 \\ \hline
\textbf{F1 (\%)} & 66.14 & 65.25 & 69.36 \\ \hline
\end{tabular}
\caption{Performance evaluation of DMD and HODMD features using SVM linear for control parameter fixed as 1000}
\label{tab:4}
\end{table}
\subsubsection{Case 3: RKS approach}
RKS approach has been used in the articles for NLP tasks [29,30,23]. In this work, we use RKS to imporve the evaluation scores as discussed previously. The RKS approach explicitly maps the embedding vectors to a dimension where the data becomes linearly separable. In that space, regularized least-square based classification (RLSC) is performed. The implementation of the RKS is taken from \cite{31,32}. The feature vectors from Google universal sentence encoder and fasttext are explicitly mapped using RKS and the results are tabulated in Table \ref{tab:5} and \ref{tab:6}. 
\begin{table}[H]
\fontsize{10pt}{10pt}
\selectfont
\centering
\begin{tabular}{|c|c|c|c|c|}
\hline
\multicolumn{5}{|c|}{\textbf{Google Embedding - RKS}} \\ \hline
\textbf{\begin{tabular}[c]{@{}c@{}}Dim $\rightarrow$\\   Score $\downarrow$\end{tabular}} & \textbf{\begin{tabular}[c]{@{}c@{}}Dim\\ =100\end{tabular}} & \textbf{\begin{tabular}[c]{@{}c@{}}Dim\\ = 200\end{tabular}} & \textbf{\begin{tabular}[c]{@{}c@{}}Dim\\ =500\end{tabular}} & \textbf{\begin{tabular}[c]{@{}c@{}}Dim\\ =1000\end{tabular}} \\ \hline
\textbf{Acc.(\%)} & 82.79 & 84.53 & 90.47 & 90.58 \\ \hline
\textbf{Prec.(\%)} & 77.38 & 81.29 & 90.31 & 90.36 \\ \hline
\textbf{Recall (\%)} & 54.17 & 57.92 & 73.75 & 74.17 \\ \hline
\textbf{F1 (\%)} & 63.73 & 67.64 & 81.19 & 81.46 \\ \hline
\end{tabular}
\caption{Performance evaluation of proposed method using Universal encoder model features}
\label{tab:5}
\end{table}

The Table \ref{tab:5} shows the classification report on the proposed RKS method taking word vectors generated by Google universal encoder model as features with dimension 512.  For this work, such vector is explicitly mapped to dimensions 100, 200, 500 and 1000 using RKS. The maximum accuracy obtained is 90.58\% for higher dimension 1000.\\

\begin{table}[H]
\fontsize{10pt}{10pt}
\selectfont
\centering
\begin{tabular}{|c|c|c|c|c|}
\hline
\multicolumn{5}{|c|}{\textbf{Fasttext Embedding - RKS}} \\ \hline
\textbf{\begin{tabular}[c]{@{}c@{}}Dim$\rightarrow$\\ Score$\downarrow$\end{tabular}} & \textbf{\begin{tabular}[c]{@{}c@{}}Dim=\\ 100\end{tabular}} & \textbf{\begin{tabular}[c]{@{}c@{}}Dim=\\ 200\end{tabular}} & \textbf{\begin{tabular}[c]{@{}c@{}}Dim=\\ 500\end{tabular}} & \textbf{\begin{tabular}[c]{@{}c@{}}Dim=\\ 1000\end{tabular}} \\ \hline
\textbf{Acc. (\%)} & 85.35 & 91.05 & 98.60 & \textbf{99.53} \\ \hline
\textbf{Prec. (\%)} & 85.19 & 90.95 & 100.00 & \textbf{99.58} \\ \hline
\textbf{Recall (\%)} & 57.50 & 75.42 & 95.00 & \textbf{98.75} \\ \hline
\textbf{F1 (\%)} & 68.66 & 82.46 & 97.44 & \textbf{99.16} \\ \hline
\end{tabular}
\caption{Performance evaluation of proposed method using Fasttext model features}
\label{tab:6}
\end{table}
Table \ref{tab:6} shows the classification report on the proposed RKS method taking word vectors generated by Fasttext model as features. For this model also, features are mapped to dimensions 100, 200, 500 and 1000. For Fasttext model, the proposed method gave a maximum accuracy of 99.53\%, which is a bench marking result when compared to the literature. This result shows the discriminating capability of the features chosen, as when mapped to higher dimensions, they become linearly separable. From Table \ref{tab:5} and \ref{tab:6} it can be observed that as the mapping dimension increases, the evaluation score also improves. This shows the effectiveness of the RKS approach to obtain competing score. The capability of the RKS approach cane be explored on large datasets. \\

\section{Conclusion}
\label{sec4}
Offensive language detection is an important task related to social media data analysis. The nature of the content can vary as its provided by different people. The current work uses the data provided in SemEval 2019 shared task A for Offensive language identification. A comparative study is provided by exploring the effectiveness of Google universal sentence encoder, Fasttext based embedding, Dynamic Mode Decomposition based features and RKS based explicit mapping approach. For the experiments, we used the machine learning methods such as SVM linear, Random Forest, Logistic regression, Navie Bayes and Regularized least-square based classification. The measures used for evaluation are accuracy, precision, recall, and f1-score. We observed that RKS approach improved the results. However, as a future work, the proposed approach cane be explored on large datasets.



\bibliographystyle{elsarticle-num}

\end{document}